\documentclass{article}
\usepackage{ijcai20}

\usepackage{times}

\usepackage{soul}
\usepackage{url}
\usepackage[hidelinks]{hyperref}
\usepackage[utf8x]{inputenc}
\usepackage[small]{caption}
\usepackage{graphicx}
\usepackage{amsmath}
\usepackage{booktabs}
\usepackage{multirow}
\usepackage{relsize}
\usepackage{tabto}
\usepackage{subcaption,graphicx}

\usepackage{latexsym} 

\title{Fed-Focal Loss for imbalanced data classification in Federated Learning}

\author{
Dipankar Sarkar$^1$\and
Ankur Narang$^1$\And
Sumit Rai$^2$\\
\affiliations
$^1$Hike Ltd.\\
$^2$IITM, Gwalior\\
\emails
dipankars@hike.in,
ankur@hike.in,
sumitrairkt@gmail.com
}

\begin{document}

\maketitle

\begin{abstract}
The Federated Learning setting has a central server coordinating the training of a model on a network of devices. One of the challenges is variable training performance when the dataset has a class imbalance. In this paper, we address this by introducing a new loss function called Fed-Focal Loss. We propose to address the class imbalance by reshaping cross-entropy loss such that it down-weights the loss assigned to well-classified examples along the lines of focal loss. Additionally, by leveraging a tunable sampling framework, we take into account selective client model contributions on the central server to further focus the detector during training and hence improve its robustness. Using a detailed experimental analysis with the VIRTUAL (Variational Federated Multi-Task Learning) approach, we demonstrate consistently superior performance in both the balanced and unbalanced scenarios for MNIST, FEMNIST, VSN and HAR benchmarks. We obtain a more than 9\% (absolute percentage) improvement in the unbalanced MNIST benchmark. We further show that our technique can be adopted across multiple Federated Learning algorithms to get improvements.

\end{abstract}

\section{Introduction}

We live in a decentralized era with large scale networks of remote devices [Lueth, 2018]. Combined with the rise of Deep Learning [Lecun \textit{et al.}, 2015], these endpoints have become a significant source of data for exciting research and applications.

In the traditional cloud model, the data is collected in centralized servers. We can then provide insights or produce inference models with the data collected. However, we see challenges of securing data centrally and the emergence of data privacy legislation like GDPR [Clusters \textit{et al.}, 2019] and the Consumer Privacy Bill of Rights [Graff \textit{et al.}, 2014].

These remote devices like mobiles and IoTs have limitations of computation and storage. That limits how complex or large the models can be. Also, we usually have a wireless network which connects them with the central server, making real-time decision systems (self-driving cars) not feasible due to propagation delays and unacceptable latency [Mao \textit{et al.}, 2017.

Collaborative paradigm [Chen \textit{et al.}, 2015] have been proposed for model training where data is sent to the edge servers to train lower DNN layers, while the server computes the resource-intensive parts. However, this approach has high communication costs, and continuous training is not feasible. Due to the data transfer from the client, there is a security concern as well. We see the usage of Differential Privacy [Abadi \textit{et al.}, 2016] in some cases, but its adoption has challenges.

Federated learning [McMahan \textit{et al.}, 2016] has become the leading paradigm to address how we can train models on private distributed data sources. A federation of devices called clients both collect data and carry out a local optimization routine. A server then coordinates the model training by receiving and sending updates to and from the clients. We have already seen its application in real-world use cases like mobile keyboard prediction [Hard \textit{et al.}, 2018].

The Federated Averaging algorithm (\textit{FedAvg}) [McMahan \textit{et al.}, 2016] and VIRTUAL (variational federated multi-task learning) [Corinzia and Buhmann, 2019] are state of the art in federated learning with non-convex models. While FedAvg performs well in a range of real-world datasets, it has not done well with strongly skewed data distributions [McMahan \textit{et al.}, 2017]. VIRTUAL overcomes this issue, with the server and client forming a Bayesian network, while inference happens using variational methods.

However, these frameworks do not tackle the global class imbalance problem as we will show in our experiments. We have seen the Astrea framework [Duan, 2019] proposed, where a rebalancing step has the clients augment their dataset in cases of imbalances. That has led to accuracy improvements. We are providing an alternative approach, where the client will not have to spend more computation to correct imbalances.

We have identified focal loss [Lin \textit{et al.}, 2018], which tackles the fore-background imbalance in image-based object detection as a starting point. In the aforementioned paper, one-stage detectors have achieved the same accuracy as two-stage detectors. The idea is to help the network focus on hard classified objects in case they are easily overwhelmed by a large number of easily classified objects.

In this paper, we establish the following contributions
\begin{itemize}
\item We introduce a new loss function called Fed-Focal Loss. By using a tunable sampling framework, our approach robustly handles unbalanced and non-IID data distribution simultaneously.
\item Detailed experimental analysis on MNIST, FEMNIST, VSN and HAR datasets with the VIRTUAL approach, demonstrates consistent superior performance of our framework for both balanced and unbalanced scenarios. Our technique leads to 9\% (absolute percentage) gain in accuracy vs prior results for the MNIST classification problem in the unbalanced scenario.
\item Using ablation study, we demonstrate the robustness of our technique in various conditions. Further, instability during training with the default cross-entropy loss gets smoothened out using our technique.
\end{itemize}

\section{Related work}
\subsection{Federated Learning}
With the proliferation of networks and increased computing power, we have seen the emergence of distributed learning where model s optimization happens in a parallel computing environment given a dataset[McDonald \textit{et al.}, 2010]. In recent years,  Federated Learning [McMahan \textit{et al.}, 2016] has introduced a new class of algorithms that can be deployed on mobile networks. In the same work, FedAvg was introduced, which has been a popular heuristic, where at each step, a sample of the clients are selected and then updated locally using SGD. An average is subsequently obtained, which is transmitted back to the clients. As stated, this approach does not do well with skewed non-IID data [Zhao \textit{et al.}, 2018].

There has been work done in embedding FL using the MTL framework using MOCHA [Smith \textit{et al.}, 2017], which builds on a few other early works. Further improvements were made by VIRTUAL [Corinzia and Buhmann, 2019], which tackles the skewed data challenges using an MTL approach.

\subsection{Class Imbalance}
In one-stage object detection methods like boosted detectors and SSD[Liu \textit{et al.}, 2016], there is a large class imbalance during training. Very few image locations contain objects. It leads to inadequate training, and easy negatives can lead to bad models. The most common solution is hard negative mining, which at times involves complex sampling schemes [Bulo \textit{et al.}, 2017]. However, Focal Loss [Lin \textit{et al.}, 2018] naturally handles this without sampling and the easy negatives overwhelming the loss and computed gradients.

We also have robust loss functions (e.g. Huber Loss [Hastie \textit{et al.}, 2008]) which reduce the contribution of outliers by down-weighting the loss of examples with large errors (hard examples). This is in contrast to Focal loss that focuses the training on a sparse set of hard examples, by down-weighing easy examples.

In FL, we have the Astrea [Duan, 2019] framework that rebalances the clients by getting them to augment their data set. After retraining with the augmented data, a mediator is created to coordinated intermediate aggregation. It selects the participants with the best data distributions, which can make useful contributions using the KL Divergence [Joyce, 2011] between the local and uniform distribution.

\begin{table*}[t]
    \centering
    \begin{tabular}{@{}llllll@{}}
        \midrule
        \textbf{Dataset} & \textbf{Number of Clients} & \textbf{Number of classes} & \textbf{Total Samples} & \textbf{Mean} & \textbf{Std} \\
        \midrule
        MNIST & 10 & 10 & 60000 & 6000 & 0\\
        Sampled-FEMNIST & 10 & 10 & 10000 & 600 & 0\\
        VSN & 23 & 2 & 68532 & 3115 & 559\\
        HAR & 30 & 6 & 15762 & 543 & 56\\
        \bottomrule
        \end{tabular} 
        \caption{The datasets used in the experiments}
    \label{Table 1}
\end{table*}

\section{Fed-Focal Loss}
The Fed-Focal Loss is designed to address an extreme imbalanced in classes during training. We introduce a crucial modification to the Focal Loss function based on the clients training performance. 
\subsection{Focal Loss}
In this section, we will describe Focal Loss.  Our experiments will show that large class imbalance will overwhelm the cross-entropy Loss. Here, we are reshaping the loss function to down weight easy examples and focus training on hard examples.
We define y $\in$ {±1} as the ground-truth class and p as the estimated probability for the class with y=1. We define the posterior probability $p_t$ as

\begin{equation}
    p= 
\begin{cases}
    1,& \text{if } y = 1\\
    1-p,& \text{if } y = -1
\end{cases}
\end{equation}

where $p=sigmoid(x)$. The binary cross-entropy loss and its deviation is

\begin{align}
\epsilon_{BCE}(p_t) = -log(p_t)\\
\frac{d\epsilon_{BCE}(p_t)}{dx} = y(p_t - 1)
\end{align}

If we train the network with BCE loss, the gradient will be dominated by vast easy classified negative samples, where there is a huge imbalance [2]. We can define Focal loss as a dynamically scaled cross-entropy loss.

\begin{gather}
\epsilon_{FL}(p_t) = -(1-p_t)^\gamma log(p_t)\\
\frac{d\epsilon_{FL}(p_t)}{dx} = y(p_t - 1)^\gamma (\gamma p_t log(p_t) + p_t - 1)
\end{gather}

The contribution from the well-classified samples (pt $\gg$ 0.5)to the loss is down-weighted. The hyperparameter of the focal loss can be used to tune the weight of different samples. As $\gamma$ increases, fewer easily classified samples contribute to the training loss. When $\gamma$ reaches 0, the focal loss degrades to become same as the BCE loss. In the following sections, all the cases with $\gamma$ = 0 represent BCE loss cases.

In previous works, techniques have balanced the losses calculated from positive and negative anchors. Alternatively,   normalized positive and negative losses by the frequency of corresponding anchors are used as well. 

However, one fundamental problem that these two previous methods cannot handle is the gradient salience of hard negative samples. A large number of easy negative anchors (pt $\gg$0.5) will overwhelm the gradients of hard negative anchors (pt $<$ 0.5). Due to the dynamic scaling with the posterior probability $p_t$, a weighted focal loss can be used to handle both the fore-background imbalance and the gradient salience of hard negative samples with the following form.
 
\begin{equation}
\epsilon_{FL}(p_t) = -\alpha(1 - p_t)^\gamma log(p_t)
\end{equation}
where $\alpha$ can be used to scale the distinct classes separately.

\subsection{Definition}

As we have seen, Focal loss leverages the dynamically changing class probability $(p_t)$ values to down-weight the well-classified samples and thereby focuses more on the challenging class samples that are significantly less in number.

In general, we rely on the validation loss during training to infer about the performance of the model. A model with lower validation loss is considered to perform better as compared to a model with higher validation loss. Once identified, the server performing the aggregation can focus on using the best-performing clients for global training.

In Federated Learning, \textit{N} clients are associated with \textit{N} datasets $D_1,\dots,D_N$, where $D_i$ denotes the dataset of client id \textit{i} is in general generated by a client dependent probability distribution function and only accessible by the respective client.

In each round of training with VIRTUAL, we sample a fixed number of \textit{K} clients from a pool of \textit{N} clients. The aggregation of weights in training thus is highly dependent on which clients were sampled in each round. These clients will refine the server model in each round.

Fed-Focal loss function takes into input the validation losses of sampled clients from the last round. As training starts, we randomly pick up \textit{K} clients from a pool of \textit{N} clients. During subsequent training rounds, we look at the validation loss from the last training round to see if some of them have improved. We expect these clients to perform better and contribute to the overall accuracy.

There can be a case of sampling bias with respect to clients. We introduce a tuning parameter called $\psi$, which ensures appropriate sampling of clients.  

\begin{gather}
0<\psi<1.0\\
c = \left \lfloor{\psi K}\right \rfloor
\end{gather}

Let $C_{i}$ be the set of $K$ clients that were selected the last training round numbered $i$. There will be a subset $V_{i}$ which will represent the clients where validation loss went down the last round. We further randomly select a subset which will contain at most $\psi K$ clients.

\begin{multline}
V_{i} = \{j \in C_{i}: \text{Where in client j} \\
\text{validation loss has decreased}\}
\end{multline}
\begin{multline}
V^{c}_{i} = \{ l \in V_{i}: \text{Where client l is a} \\
\text{random selection}\}
\end{multline}

We will now have to select the right clients $C_{i+1}$ for the next round, and correspondingly compute the losses using Focal Loss. When $i=0$, which is the first training cycle $C_{0}$ will be randomly selected. 

\begin{gather}
k = \vert V^{c}_{i}\\
R_{i+1} = \text{Random selection of any} (K-k) \text{clients} \\
C_{i+1} = V^{c}_{i} \cup R_{i+1}
\end{gather}

The server now has its focus on clients that are performing well. If $\psi$ is tuned appropriately, as shown in the experiments, it will smoothen the training and improve accuracy.

\begin{table*}[t]
 \centering
 \begin{tabular}{@{}lllll@{}}
     \midrule
 \textbf{Method} & \textbf{MNIST} & \textbf{Sampled-FEMNIST} & \textbf{VSN} & \textbf{HAR} \\
 \midrule
 Global & 0.9678$\pm$0.0007 & 0.8326$\pm$0.008 & 0.926$\pm$0.004 & 0.797$\pm$0.006 \\
 Local & 0.9511$\pm$0.0020 & 0.8299$\pm$0.005 & 0.960$\pm$0.003 & 0.940$\pm$0.001 \\
 FedAvg & 0.9675$\pm$0.0004 & 0.8319$\pm$0.003 & 0.916$\pm$0.007 & 0.944$\pm$0.004\\
 Virtual(CE) & 0.9666$\pm$0.0017 & 0.8147$\pm$0.01 & 0.9340$\pm$0.002 & 0.9680$\pm$0.001\\
 Virtual(FFL) & \textbf{0.9690$\pm$0.0010} & \textbf{0.8315$\pm$0.01} & \textbf{0.9412$\pm$0.002} & \textbf{0.9692$\pm$0.002}\\ 
 \bottomrule
 \end{tabular}
 \caption{Multi-task average test accuracy on balanced dataset}
 \label{Table 2}
\end{table*}

\begin{table*}[t]
 \centering
 \begin{tabular}{@{}lllll@{}}
     \midrule
 \textbf{Loss function} & \textbf{MNIST} & \textbf{FEMNIST} & \textbf{VSN} & \textbf{HAR} \\
 \midrule
 Cross-entropy & 0.6660$\pm$0.0015 & 0.6750$\pm$0.01 & 0.7981$\pm$0.002 & 0.8205$\pm$0.001\\
 Fed-Focal & \textbf{0.7591$\pm$0.0012} & \textbf{0.7555$\pm$0.01} & \textbf{0.8249$\pm$0.002 } & \textbf{0.8501$\pm$0.004}\\ 
 \bottomrule
 \end{tabular}
 \caption{Test accuracy using VIRTUAL on unbalanced dataset}
 \label{Table 3}
\end{table*}

\section{Experiments}

In this section, we present the empirical evaluation of the performance of Fed-Focal Loss on various standard federated datasets. We have used the VIRTUAL codebase, as it is a state of the art FL algorithm and extended it with our loss function. 

\subsection{Datasets}

We have listed the nature of the datasets in Table 1.

\begin{itemize}
\item \textbf{MNIST:}  The dataset contains a total of 70,000 small square 28x28 pixels grayscale images of handwritten single digits between 0 and 9. The subdivision of dataset is 60,000 training samples and 10,000 testing samples. We federated the dataset amongst the clients to simulate the real world MTL.

\item \textbf{Sampled-FEMNIST:} This dataset consists of a federated version of the EMNIST dataset, maintained by the LEAF project. We sampled 10,000 training and 1000 testing samples to simulate the real world MTL. We selected this dataset with the specified sampling volume to simulate the MTL in a real-world environment where the dataset is scarce.

\item \textbf{Vehicle Sensors Network (VSN):} A network of 23 different sensors (including seismic, acoustic and passive infra-red sensors) are placed around a road segment in order to classify vehicles driving by them [Duarte and Hu, 2004]. The raw signal is processed in the original paper into 50 acoustic and 50 seismic features. We consider every sensor as a client and perform binary classification of amphibious assault vehicles and dragon wagon vehicles.

\item \textbf{Human Activity Recognition (HAR):} Recordings of 30 subjects performing daily activities are collected using a waist-mounted smartphone with inertial sensors. The raw signal is divided into windows and processed into a 561-length vector [D Anguita \textit{et al.}, 2013]. Every individual corresponds to a different client, and we perform a classification of 6 different activities (e.g. sitting, walking).

\end{itemize}

\subsection{Settings}

All networks employed are multilayer perceptrons (MLP) with two hidden dense Flipout layers with 100 units, and ReLU activation functions and final layer is of 10 units and softmax activation function. The Monte Carlo estimate of the gradient is performed in all experiments using a ratio of 0.10 clients in each round (unless specified otherwise).  

We use the VIRTUAL as a testbed for comparing the loss function, due to its ability to handle skewed data by design. Given that it has outperformed FedAvg, we see it as a useful framework for our work.

In all the experiments, we evaluate Fed-Focal Loss (FFL) while incrementally training the clients in a fixed order of the sampled clients with one refinement per client. We used $\alpha = 1.0$ and $\gamma = 2.0$ for all experiments (unless specified) and varied the value of $\psi$.We used $\psi = 0.8$ for Sampled FEMNIST and MNIST dataset and $\psi = 0.6$ for VSN and HAR dataset. We performed the experiments with balanced and unbalanced sets of the mentioned datasets. This is to compare the usage of Cross-Entropy Loss and our Fed-Focal Loss.

\begin{itemize}

\item \textbf{Balanced Dataset:} The balanced dataset consists of all the images of the corresponding dataset where all classes have the same class ratio.

\item \textbf{Unbalanced Dataset:} The unbalanced dataset consists of four classes, namely the digits 0,1,2,9 being unbalanced in the ratio of 1:100 in case of MNIST and 1:10 in case of Sampled-FEMNIST. We unbalance single class in VSN and HAR dataset in the ratio of $1:20$.

\end{itemize}

\subsection{Results}

\subsubsection{Balanced}

We used the balanced version of the datasets to get a baseline of the performance. In Table 2, we see that VIRTUAL with FFL does consistently better than the CE version. Here we measure the average accuracy over all classes on the test set and report the mean and standard deviations over the runs. 

We can see that the performance depends strongly on the volume distribution and heterogeneity of the dataset under consideration. In particular, the Global baseline is among the top-performing methods only on the MNIST dataset, that has been generated in the IID fashion.

The VSN dataset depicts strongly non-IID scenarios that have significantly dissimilar feature spaces among clients. VSN encompasses a broad spectrum of different sensors. The Sampled-FMNIST dataset depicts the problem of few datapoints with clients (70-100 samples per client). 

In these scenarios, the performance of both Global and FedAvg degrades while Local models enjoy high performances, being tuned to the specific data distribution. We can see that VIRTUAL, when used with Fed-Focal Loss, maintains the top performance in the whole spectrum of federated scenarios, being on par with Global and FedAvg on IID datasets, and with Local models on strongly non-IID settings. 

It also outperforms other methods on Sampled FEMNIST and HAR, that are datasets that best represent the multi-task learning setting, as they encompass different users gathering data in very similar but distinct conditions.

\begin{table*}[t]
    \centering
    \begin{tabular}{@{}llllll@{}}
    \midrule 
     \textbf{Loss} & \textbf{Clients (ratio)} & \textbf{Balanced MNIST} & \textbf{Unbalanced MNIST} & \textbf{Balanced FEMNIST} & \textbf{Unbalanced FEMNIST} \\ 
    \midrule 
    Cross-entropy & 0.05 & 0.9507$\pm$0.0010 & 0.5574$\pm$0.0015 & 0.7614$\pm$0.015 & 0.5159$\pm$0.010 \\ 
    Fed-Focal & 0.05 & 0.9590$\pm$0.0009 & 0.5635$\pm$0.0012 & 0.7814$\pm$0.015 & 0.5448$\pm$0.012 \\ 
    Cross-entropy & 0.10 & 0.9687$\pm$0.0011 & 0.6660$\pm$0.0014 & 0.8147$\pm$0.010 & 0.6750$\pm$0.010 \\ 
    Fed-Focal & 0.10 & 0.9707$\pm$0.0070 & 0.7571$\pm$0.0015 & 0.8315$\pm$0.010 & 0.7555$\pm$0.010 \\ 
    Cross-entropy & 0.15 & 0.9631$\pm$0.0017 & 0.7570$\pm$0.0019 & 0.7751$\pm$0.013 & 0.7321$\pm$0.012 \\ 
    Fed-Focal & 0.15 & 0.9666$\pm$0.0030 & 0.7710$\pm$0.0032 & 0.7911$\pm$0.015 & 0.7458$\pm$0.014 \\ 
    \bottomrule 
    \end{tabular}
    \caption{Varying the number of clients sampled}
    \label{Table 6}
\end{table*} 

\subsubsection{Unbalanced}

Given the baselines we have set in the balanced scenario, we have done experiments with unbalanced MNIST. In Table 3, we observe the performance of FFL to classify hard samples ($p_t \ll$ 0.5) under single class and multi-class imbalance when compared with the cross-entropy.

\begin{table}[h]
    \centering
    \begin{tabular}{@{}lllll@{}}
        \midrule
        $\gamma$ & Accuracy \\ 
        \midrule 
        1.0 & 0.7486$\pm$0.0001 \\ 

        2.0 & 0.7840$\pm$0.0015 \\ 

        3.0 & 0.6786$\pm$0.0040 \\ 

        4.0 & 0.6736$\pm$0.0043 \\ 

        5.0 & 0.6220$\pm$0.0052 \\ 
        \bottomrule 
    \end{tabular}
    \caption{Test Accuracy on Unbalanced MNIST at $\psi=0.0$ and $\alpha=1.0$}
    \label{Table 4}
\end{table}

We study the variation of test accuracy with the tuning parameter $\gamma$ of Fed Focal Loss. We can observe in Table 5 that smaller values of $\gamma$ are suitable while larger values of $\gamma$ lead to worse performance. $\gamma$ = 1.0 and 2.0 deliver very good test set accuracies as compared to $\gamma$ = 3.0, 4.0 and 5.0.

\begin{table}[h]
    \centering
    \begin{tabular}{@{}lllll@{}}
    \midrule 
    $\psi$ & Accuracy \\ 
    \midrule
    0.0 & 0.7486$\pm$0.0001 \\ 

    0.2 & 0.7497$\pm$0.0001 \\ 

    0.6 & 0.7504$\pm$0.0002 \\ 

    0.8 & 0.7591$\pm$0.0010 \\ 

    1.0 & 0.7473$\pm$0.0030 \\ 
    \bottomrule
    \end{tabular} 
    \caption{Test Accuracy on Unbalanced MNIST at $\gamma=2.0$ and $\alpha=1.0$}
    \label{Table 5}
\end{table}

We also study the variation of test accuracy with the tuning parameter $\psi$ of Fed Focal Loss. $\psi$ = 0.0  is when no particular focus is given to the client, while $\psi$ = 1.0  is when we focus only on all clients that have improved their performance in the previous round. In Table 6, we can observe that $\psi$ = 0.6 and 0.8 shows a significant improvement in the performance. Although we cannot generalize a specific tuned value as it strongly depends on the distribution of validation dataset, we found the best performance with $\psi$ = 0.8 and 0.6.

We can observe that in the class imbalance situation, the performance of FFL based VIRTUAL surpasses the benchmark accuracy of unbalanced MNIST, Sampled-FEMNIST, VSN and HAR. It is observed than the performance strongly depends on the distribution of a dataset. In the class imbalance situation, few clients may have zero samples of unbalanced class. 

This condition is depicted strongly by the Sampled FEMNIST. The classic MNIST dataset depicts the IID scenario, and non-IID scenario is depicted by VSN dataset. Despite the highly skewed dataset distribution, the Fed Focal Loss performs better than cross-entropy.

\subsection{Ablation study}
 
We perform an ablation study to validate the efficacy of our Fed-Focal Loss. These experiments are conducted on MNIST and Sampled-FEMNIST. 

\subsubsection{Client sampling}

This study focused on how changing the number of clients to sample each round impacted the performance. We can see that in Table 4, the performance of both methods decreases at the sampling ratio of 0.05 or 0.15 clients, and the best results are obtained with a ratio of 0.10 clients. 

Moreover, class accuracy of unbalanced class decreases drastically in case of the sampling ratio 0.05 as compared to 0.10 and 0.15. Despite varying the sampling frequency, FFL maintains the top performance under both the balanced and unbalanced case.

\subsubsection{Noisy MNIST}
The Classic MNIST dataset consists of black and white images with pixel values nearer to 0 depicting a black shade and pixel values nearer to 255 means a lighter shade. Real-world datasets are affected with noise and hence to simulate the real-world scenario, we introduce some noise in MNIST dataset. 

We choose additive Gaussian noise to manipulate the pixel values from their default values. This manipulation can be controlled by two parameters of Gaussian distribution, mean ( $\mu$ ) and standard deviation ( $\sigma$ ). We select $\mu = 10$ and $\sigma = 5$ to generate a Gaussian noise to manipulate each pixel value and perform the addition operation to achieve a Noisy MNIST dataset.

\begin{table}[h]
    \centering
    \begin{tabular}{@{}lllll@{}}
    \midrule 
    \textbf{Loss} & \textbf{Balanced} & \textbf{Unbalanced} \\ 
    \midrule 
    Cross-entropy & 0.9381$\pm$0.0010 & 0.6202$\pm$0.0017 \\ 
    Fed-Focal  & \textbf{0.9552$\pm$0.0010} & \textbf{0.6692$\pm$0.0017} \\ 
    \bottomrule
    \end{tabular} 
    \caption{Using Noisy MNIST with 0.10 ratio of clients}
    \label{Table 7}
\end{table} 

The comparisons are shown in Table 7, where the proposed Fed-Focal Loss gets the best estimation.

\section{Observations}

\subsection{Extreme imbalance}
When the class imbalance is more than 1:50 in VSN and HAR dataset, it becomes difficult for both the loss functions to correctly classify the unbalanced classes. Although in some cases it was observed that 2-5 samples were correctly classified by Fed-Focal Loss and none of the samples was classified correctly by Cross-entropy Loss. Similar behaviour was seen with MNIST at 1:500 imbalance and Sampled FEMNIST at 1:100 and 1:200 class imbalance ratios.

\subsection{Smoother convergence}

During training on the classic MNIST balanced dataset, Figure 1 depicts the accuracy comparison of Cross-entropy with Fed-Focal Loss. We observe that without using Fed Focal Loss ( $\psi$ = 0.0, $\gamma$ = 0.0, $\alpha$ = 1.0 ) the graph is unstable as compared to using Fed-Focal Loss with ( $\psi$ = 0.8, $\gamma$ = 2.0, $\alpha$ = 1.0 ).

\begin{figure}[!h]
\centering
\includegraphics[scale=0.5]{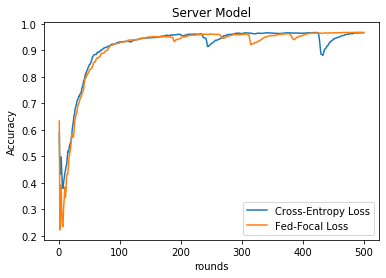}
\caption{MNIST training comparison - Accuracy}
\label{Figure 1}
\end{figure}

\begin{figure}[!h]
\centering
\includegraphics[scale=0.5]{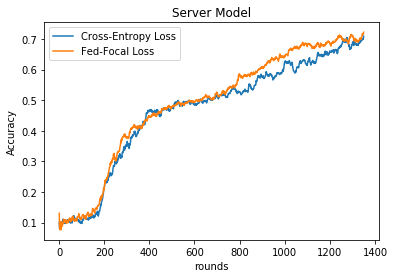}
\caption{FEMNIST training comparison - Accuracy}
\label{Figure 2}
\end{figure}

In Figure 2, we see that Fed-Focal Loss and cross-entropy follow a similar accuracy curve until 800 rounds after which Fed-Focal Loss seems to perform better.

\begin{figure}[!h]
\centering
\includegraphics[scale=0.5]{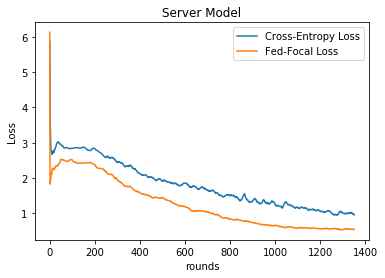}
\caption{FEMNIST training comparison - Loss}
\label{Figure 3}
\end{figure}

Similarly, when we train on unbalanced Sampled-FEMNIST dataset, we observe smoother convergence in Figure 3.

\section{Conclusions \& Future Work}

In this paper, we extended the Focal Loss function used in image detectors to Federated Learning along with tunable sampling framework for solving the class imbalance problem. By getting the loss to focus on hard-examples and better-performing clients feedback, we have shown significant improvements inaccuracy.

Experimental analysis with VIRTUAL approach on MNIST, FEMNIST, VSN and HAR benchmarks, demonstrates clear accuracy improvements over cross-entropy based loss function. The ablation study further shows the robustness of FFL in various conditions of clients behaviour and dataset noise.

We further observe training stability using FFL during the experiments. We believe the function will help smoother convergence in cases where the data has challenges. In extreme cases of class imbalance in data, it does better than Cross- entropy.

We clearly see the application of this loss function in popular FedAvg-based frameworks like \textit{PySft} [Ryffel \textit{et al.}, 2018] to gain similar improvements in more complex network types as well. The Federated Learning paradigm has the additional benefits of monitoring client behaviour, and it will be incorporated into the various aspects to gain training stability and accuracy going forward.

\section*{Acknowledgements}

We would like to thank the VIRTUAL paper authors who shared their current codebase with us and answered our questions about their results.

\section*{References}

\hangafter=1
\hangindent=1em
\noindent [Lueth, 2018] K. L. Lueth. State of the IoT 2018: Number of IoT devices now at 7 Billion market accelerating. \textit{IOT Analytics}, 2018.\\

\hangafter=1
\hangindent=1em
\noindent [LeCun \textit{et al.}, 2018] LeCun, Y. Bengio, and G. Hinton. Deep learning. \textit{Nature}, vol. 521,no. 7553, p. 436, 2015. \\

\hangafter=1
\hangindent=1em
\noindent [Clusters \textit{et al.}, 2019] B. Custers, A. Sears, F. Dechesne, I. Georgieva, T. Tani, and S. van derHof. EU Personal Data Protection in Policy and Practice.\textit{Springer},2019. \\

\hangafter=1
\hangindent=1em
\noindent [Graff \textit{et al.}, 2014] B. M.Gaff, H.E.Sussman, and J.Geetter.Privacy and big data. \textit{Computer}, vol. 47, no. 6, pp. 7–9, 2014.\\

\hangafter=1
\hangindent=1em
\noindent [Mao \textit{et al.}, 2017] Y. Mao, C. You, J.Zhang, K.Huang, and  K. B. Letaief. A survey on  mobile edge  computing: The communication  perspective.\textit{IEEECommunications  Surveys \& Tutorials}, vol.19, no. 4, pp. 2322–2358,2017.\\

\hangafter=1
\hangindent=1em
\noindent  [Chen \textit{et al.}, 2015] X. Chen, L. Jiao, W. Li, and X. Fu. Efficient multi-user computation offloading for mobile-edge cloud computing. \textit{IEEE/ACM Transactions on Networking}, vol. 24, no. 5, pp. 2795–2808, 2015.\\
 
\hangafter=1
\hangindent=1em
\noindent [Abadi \textit{et al.}, 2016] M.  Abadi,  A.  Chu,  I.  Goodfellow,  H.  B.  McMahan,  I.  Mironov, K.  Talwar, and  L.  Zhang. Deep learning with differential privacy. In \textit{Proceedings  of  the  2016  ACM  SIGSAC  Conference  on  Computer and Communications Security.}, ACM pp. 308–318, 2016.\\

\hangafter=1
\hangindent=1em
\noindent [McMahan \textit{et al.}, 2016] H. B. McMahan, E. Moore, D. Ramage, and B. A. y Arcas. "Federated learning of deep networks using model averaging", 2016.\\

\hangafter=1
\hangindent=1em
\noindent [Hard \textit{et al.}, 2018] A. Hard, K. Rao, R. Mathews, F. Beaufays, S. Augenstein, H. Eichner, C. Kiddon, and D. Ramage. Federated learning for mobile keyboard prediction. \textit{arXiv preprint} arXiv:1811.03604, 2018.\\

\hangafter=1
\hangindent=1em
\noindent [Corinzia and Buhmann, 2019] Luca Corinzia, Joachim M. Buhmann. Variational Federated Multi-Task Learning.\textit{arXiv preprint} arXiv:1906.06268, 2019.\\

\hangafter=1
\hangindent=1em
\noindent [McMahan \textit{et al.}, 2017] Brendan McMahan et al. Communication-Efficient Learning of Deep Networks from Decentralized Data. In \textit{Artificial Intelligence and Statistics} pp. 1273–1282, 2017.\\

\hangafter=1
\hangindent=1em
\noindent [Duan, 2019] M.  Duan. Astraea:  Self-balancing  federated  learning  for  improving classification  accuracy  of  mobile  deep  learning  applications. \textit{arXiv preprint} arXiv:1907.01132, 2019.\\

\hangafter=1
\hangindent=1em
\noindent [Lin \textit{et al.}, 2018] T. Lin, P. Goyal, R. Girshick, K. He, and P. Dollar. Focal loss for dense object detection.\textit{IEEE Transactions on Pattern Analysis and Machine Intelligence}, pp. 1–1, 2018.\\

\hangafter=1
\hangindent=1em
\noindent [McDonald \textit{et al.}, 2010] Ryan McDonald, Keith Hall, and Gideon Mann.Distributed training strategies for the structured perceptron. In \textit{Human Language Technologies: The 2010 Annual Conference of the North American Chapter of the Association for Computational Linguistics. Association for Computational Linguistics}, pp. 456–464, 2010.\\

\hangafter=1
\hangindent=1em
\noindent [Zhao \textit{et al.}, 2018] Yue Zhao et al. Federated learning with non-iid data. \textit{arXiv preprint} arXiv:1806.00582, 2018.\\

\hangafter=1
\hangindent=1em
\noindent [Smith \textit{et al.}, 2017] Virginia Smith et al. Federated multi-task learning. \textit{ Advances in Neural Information Processing Systems}, pp. 4424–4434, 2017.\\

\hangafter=1
\hangindent=1em
\noindent [Liu \textit{et al.}, 2016] W. Liu, D. Anguelov, D. Erhan, C. Szegedy, and S. Reed. SSD: Single shot multibox detector. In \textit{ECCV}, 2016.\\

\hangafter=1
\hangindent=1em
\noindent [Bulo \textit{et al.}, 2017] S. R. Bulo, G. Neuhold, and P. Kontschieder. Loss max-pooling for semantic image segmentation. In \textit{CVPR}, 2017.\\

\hangafter=1
\hangindent=1em
\noindent [Hastie \textit{et al.}, 2008] T. Hastie, R. Tibshirani, and J. Friedman. The elements of statistical learning. \textit{Springer series in statistics Springer, Berlin}, 2008.\\

\hangafter=1
\hangindent=1em
\noindent [Joyce, 2011] J. M. Joyce. Kullback-leibler divergence. \textit{International encyclopedia of statistical science}, pp. 720–722, 2011.\\

\hangafter=1
\hangindent=1em
\noindent [Duarte and Hu, 2004] Marco F Duarte and Yu Hen Hu.Vehicle classification in distributed sensor networks. \textit{Journal of Parallel and Distributed Computing} 64.7, pp. 826–838, 2004.\\

\hangafter=1
\hangindent=1em
\noindent [D Anguita \textit{et al.}, 2013] D Anguita et al. A Public Domain Dataset for Human Activity Recognition using Smartphones .\textit{21th European Symposium on Artificial Neural Networks, Computational Intelligence and Machine Learning (ESANN)}, CIACO, pp. 437–442, 2013.\\

\hangafter=1
\hangindent=1em
\noindent [Ryffel \textit{et al.}, 2018] T.  Ryffel,  A.  Trask,  M.  Dahl,  B.  Wagner,  J.  Mancuso,  D.  Rueckert, and J. Passerat-Palmbach. A generic framework for privacy-preserving deep learning.\textit{arXiv preprint} arXiv:1811.04017, 2018

\end{document}